\documentclass[letterpaper]{article} 
\usepackage{aaai23}  
\usepackage{times}  
\usepackage{helvet}  
\usepackage{courier}  
\usepackage[hyphens]{url}  
\usepackage{graphicx} 
\usepackage{natbib}  
\usepackage{caption} 
\usepackage{algorithm}
\usepackage{algorithmic}

\usepackage{nicefrac}
\usepackage{amsfonts}
\usepackage{amsmath}
\usepackage{booktabs}
\usepackage{subcaption}
\usepackage{adjustbox}

\usepackage{newfloat}
\usepackage{listings}
\DeclareCaptionStyle{ruled}{labelfont=normalfont,labelsep=colon,strut=off} 
\floatstyle{ruled}
\newfloat{listing}{tb}{lst}{}
\floatname{listing}{Listing}
\title{Few-Shot Out of Domain Intent Detection with Covariance Corrected
  Mahalanobis Distance}
\author {
    Jayasimha Talur,
    Oleg Smirnov,
    Paul Missault
}
\affiliations {
  Amazon\\
  \{talurj, osmirnov, pmissaul\}@amazon.com
}

\begin{document}

\maketitle

\begin{abstract}
Conversational agents like chat bots and voice assistants are trained to
understand and respond to user intents. On encountering an utterance with an
intent different from the ones they have been trained on, these agents are
expected to classify the intent as ``unknown'' or ``out of domain''. This
problem is known as out of domain (OOD) intent
detection. \citet{Podolskiy2021RevisitingMD}, showed that Mahalanobis distance
can be used effectively for identifying OOD intents, outperforming competing
approaches. However, their method fails to outperform the baselines in the
practically important few-shot setting. In this paper we analyze the reason
for low performance and propose a covariance corrected Mahalanobis distance
for detecting out-of-domain intents.
\end{abstract}

\section{Introduction}
\label{sec:introduction}
Intent classification is a key component of natural language understanding
systems, such as voice assistants and chat bots. Recent advances in those
systems are overwhelmingly contributed by deep learning techniques, that can
learn meaningful feature representations with a minimum amount of
hand-crafting~\citep{surveyIntent}. Chat bots typically follow the
intent-response pattern, where there is a fixed or context-aware mapping
between predicted intents and the responses. In addition to providing the
confidence score for intent, intent detection models are also expected to
produce an OOD score, that measures the likelihood of an utterance being
out-of-domain. Typically, when an utterance is deemed to OOD, a fallback
mechanism is triggered to either ask a clarifying question or respond with ``I
don't know''. OOD detection can be modeled as a binary classification task,
where we are interested in classifying the utterance in out-of-domain and
in-domain (IND) categories. For good user experience and user trust it is
important to achieve a strong trade-off between precision and recall of the
OOD classifier.

\begin{figure}[t]
	   \centering
		\includegraphics[width=0.95\columnwidth,keepaspectratio]{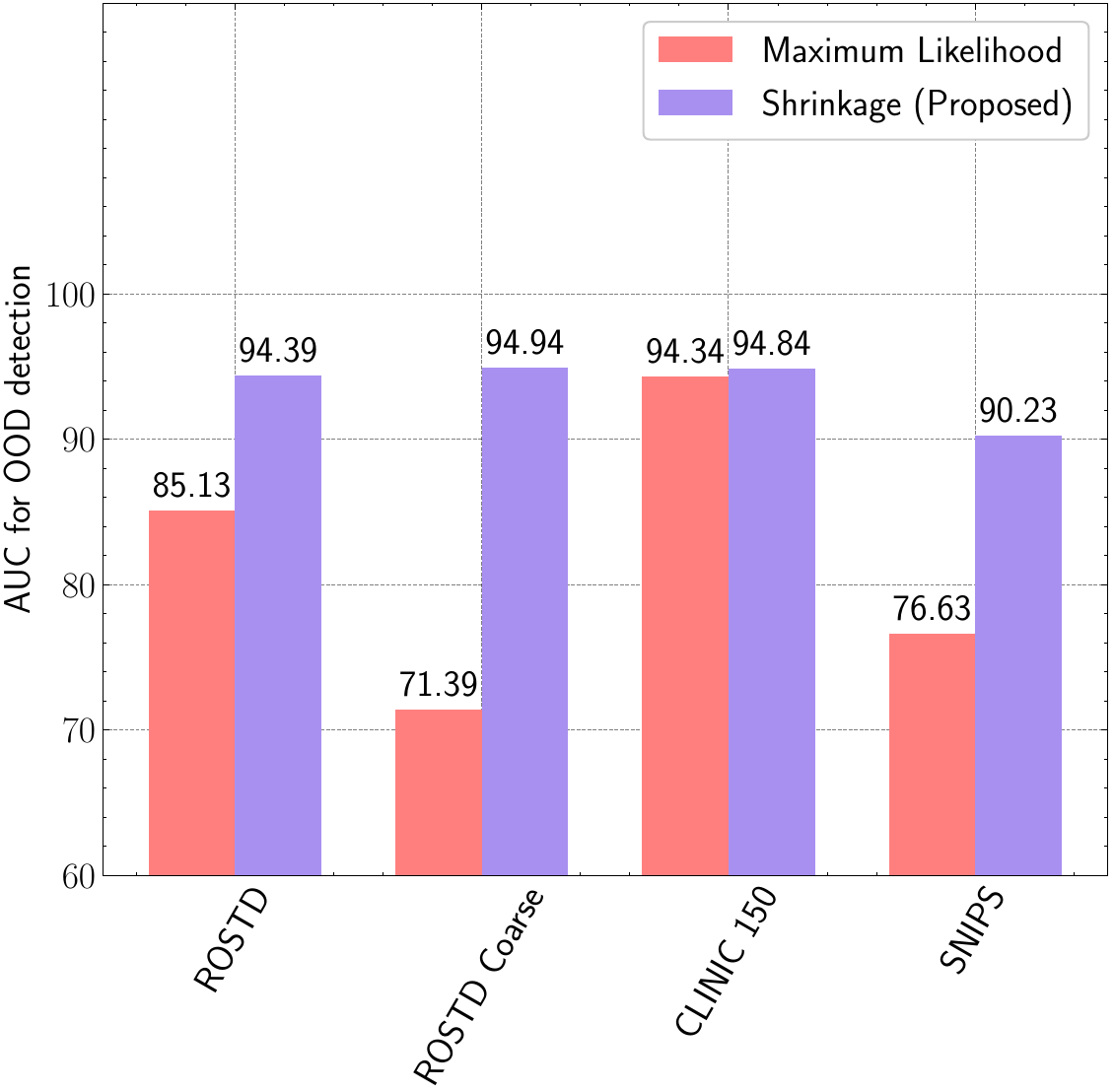}
		\caption{Performance of OOD detection for various intent
          classification datasets when Mahalanobis distance is calculated
          using Shrinkage (our proposal) and Maximum Likelihood estimators in 5-shot
          setting.}
		\label{fig:baseline_comparison}
\end{figure}

Recently many methods have been proposed to detect OOD
intents~\citep{Podolskiy2021RevisitingMD,goldpaper,pnPOOD}. However, these
methods require access to a large training dataset, either with out-of-domain
examples, or a large unlabeled corpus. In industrial settings it is
unreasonable to expect any of those assumptions to be met. For example, when
bootstrapping a domain-specific chat bot, there is no access to a large
training dataset of out-of-domain utterances, since the chat bot has not been
in production yet. It is also difficult to obtain a large corpus of unlabeled
\emph{domain-specific} examples.

Specifically, for conversational agents provided as a service, such as Amazon
Lex and Google Dialogflow, customers can extend the agent's capabilities by
uploading custom utterances and intents datasets. In those settings both the
training and validation data are scarce.

\citet{Podolskiy2021RevisitingMD} showed that Mahalanobis distance computed on
RoBERTa~\citep{roberta} embeddings outperforms baseline methods for OOD
detection without using any additional data. Unfortunately, Mahalanobis
distance performs poorly in low resource settings~\citep{NoClearWinner}. In
this paper, we identify the reason for the poor performance and propose a new
method, that performs OOD intent detection in low resource settings.

\section{Preliminaries and Related Work}
\label{sec:preliminaries}

The OOD detection task is to classify a test data point into OOD and IND
categories. We can broadly classify the approaches into two buckets:

\begin{itemize}
	\item Data-centric: these methods use additional OOD data to learn
	representation that can better separate OOD from IND examples. Additional
	OOD data is obtained by either sampling from a large
	corpus~\citep{Hendrycks2019DeepAD}, by using a language model to generate
	sentences~\citep{pnPOOD}, or by mining or filtering examples using sentence
	similarity models~\citep{goldpaper}.

	\item Score-based: these methods compute a \emph{score} to decide between
	the IND and OOD classes. The score can be computed from the
	features~\citep{MahalanobisOrig}, model logits~\citep{EnergyOOD,ODIN}, or
	the norm in the gradients space~\citep{GradNorm}.
\end{itemize}

In this work we focus on score-based methods, which don't require additional
OOD data that makes them attractive for industrial applications. In
the score-based methodology, given a test sample $x$ and a decision threshold
${T}$, we are interested in constructing a score function $G: x \rightarrow
\mathbb{R}$, such that $G(x) >= {T}$ implies that $x$ is OOD and $G(x) < {T}$
implies that $x$ is IND.

\textbf{Mahalanobis distance}: Let $F \in \mathbb{R}^{n \times d}$ denote $n$
points, each represented by $d$ dimensional features, and $y \in [1, C]$ the
corresponding labels in the set of $C$ classes. For a test feature $x \in R^{d
  \times 1}$, the Mahalanobis distance for OOD detection is defined by

\begin{equation}
	d(x) = \underset{c \in [1, C]}{\min}(x - \mu_c)^T \Sigma^{-1} (x - \mu_c)
\end{equation}

where  $\mu_c \in R^{d  \times 1}$ is the empirical mean of features of the
corresponding class, and $\Sigma \in R^{d \times d}$ is the feature covariance
matrix. Mahalanobis-based OOD detection method uses a score function $G(x) =
d(x)$.

Besides  OOD detection, Mahalanobis distance has been used to perform
pattern recognition~\citep{mahalanobisSurvey}, anomaly
detection~\citep{anomalydetection} and detecting adversarial
examples~\citep{MahalanobisOrig}. Mahalanobis distance is known to
performs well for sufficiently large dataset sizes. However, its performance
degrades rapidly in low resource settings~\citep{NoClearWinner}.

To the best of our knowledge, there are no score-based methods specifically
designed for few-shot out-of-domain intent detection.

\section{Methodology}
\label{sec:methodology}

\begin{table*}[]
	\resizebox{\textwidth}{!}{%
		\begin{tabular}{lllllll}
			\hline
			Training mode & Covariance data points & AUC & PR ROC$_{ood\_neg}$& PR ROC $_{ood\_pos}$& FPR$_{ood\_neg}$& FPR$_{ood\_pos}$\\ \hline
			5-shot  & 60  & 84.98$\pm$4.15 & 94.75$\pm$1.53 & 58.55$\pm$8.68 & 75.42$\pm$10.41 & 37.95$\pm$7.57 \\
			\addlinespace[0.1cm]
			5-shot  & 30K & 98.92$\pm$1.43 & 99.63$\pm$0.49 & 96.62$\pm$4.68 & 4.92$\pm$9.87   & 4.07$\pm$5.06  \\
			\addlinespace[0.1cm]
			Full            & 30K & 99.8$\pm$0.1  & -           & 99.5$\pm$0.3  & 1.0$\pm$0.5    & 0.5$\pm$0.4   \\ \hline
		\end{tabular}%
	}
	\caption{Mahalanobis OOD detection performance on the ROSTD dataset with
		respect to the number of examples for covariance estimation and training
		regimes. Performance figures for the full dataset are taken
		from~\citet{Podolskiy2021RevisitingMD}.}
	\label{tab:all_data}
\end{table*}

\begin{figure}[tp]
	\centering
		\includegraphics[width=0.9\columnwidth, keepaspectratio]{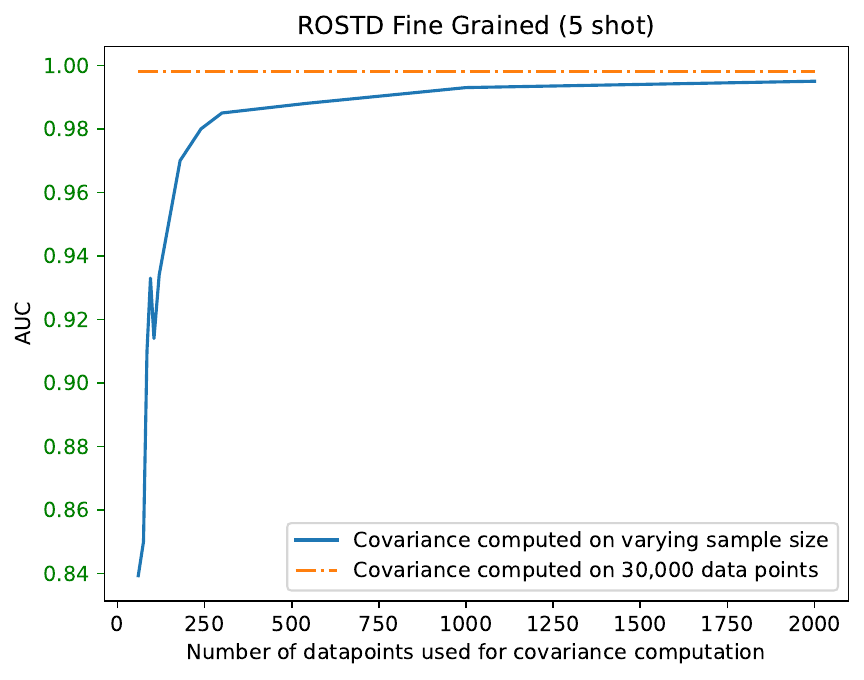}
		\caption{Performance of Mahalanobis-based OOD detection on the ROSTD
			dataset in 5-shot settings, as a function of the number of data points
			used for covariance computation. There is a sharp improvement in AUC
			with up to 400 data points, and only a modest improvement afterwards.}
		\label{fig:main_exp}
\end{figure}
Before describing our method, we analyze why Mahalanobis distance performs
worse when the training dataset size is small. \citet{NoClearWinner}
conjectured that poor covariance estimate in low sample settings leads to bad
estimate of Mahalanobis distance. The reasoning is that the rank of a $d
\times d$ covariance matrix computed on $n$ points in $d$ dimensions is
bounded from above by $\min(n-1, d)$, since $n \ll d$ in few-shot settings,
the covariance matrix becomes singular. Mahalanobis distance uses the inverse
of covariance, hence estimating a ``best fit'' solution with pseudoinverse
calculation may negatively affect the performance.

To test this hypotheses, we perform a simple experiment:
\begin{enumerate}
\item Fine-tune the RoBERTa~\citep{roberta} base model on the ROSTD intent
  classification dataset in a 5-shot setting. The ROSTD dataset is discussed
  in Section~\ref{sec:datasets}, and the fine-tuning procedure is described in
  Section~\ref{sec:train_procedure}.
\item Extract features from the last hidden layer for the full training
  dataset ($\sim${30K} data points). Note that this full dataset is
  unavailable in practical cases, since we only observe 5 data points.
\item Compute the covariance matrices using various sample sizes of training
  features extracted in the previous step.
\item Use the Mahalanobis method for detecting OOD samples.
\end{enumerate}

Table~\ref{tab:all_data} summarizes the performance of OOD detection with
respect to the metrics discussed in Section~\ref{sec:datasets}, where the
covariance matrix is computed in small and large data regimes. We observe that
the Mahalanobis distance-based OOD method trained with only 5 samples per
class performs poorly when the covariance is computed using 60 data
points. However, it achieves competitive performance provided that the
covariance matrix was computed using 30K data points. As expected, the best
performance is obtained when the covariance computation and model training was
performed on the full dataset.

Figure~\ref{fig:main_exp} depicts the OOD detection performance, when the
number of data points used for covariance computation is varied. We observe a
sharp improvement in AUC for up to 400 data points, and only a modest
improvement afterwards. This experiment empirically confirms two
assumptions. Firstly, the features extracted from a model fine-tuned on only a
handful of samples still have sufficient representational power to separate
IND and OOD categories. Secondly, a non-invertible covariance matrix
contributes to the poor performance in few-shot setting.

To overcome this, we propose to use robust covariance estimators for
covariance computation. This is motivated by the fact that robust approaches
provide a better estimate of the covariance matrix when $n \ll d$, compared to
the baseline Maximum Likelihood estimator (MLE) in the standard Mahalanobis
distance. Intuitively, this is achieved by incorporating various prior beliefs
about the structure of the features space (e.g. shape of the
clusters). However, in practice different assumptions lead to differences in
the downstream performance. Below, we briefly review covariance estimators
with their corresponding closed-form expressions listed in
Table~\ref{tab:estimators}.

\textbf{Maximum Likelihood estimator}: MLE $\Sigma_m$ is conventionally used
for computing the Mahalanobis distance. When $\Sigma_m$ is not invertible,
$\Sigma_m^{-1}$ can be estimated with a pseudoinverse.

\textbf{Van Ness estimator}: the diagonal elements of a covariance matrix
represent the variance of individual features, that is typically non-zero for
all elements. Van Ness estimator~\citep{vanness} only retains the diagonal
elements $\Sigma_{\text{Van Ness}}$ and sets non-diagonal elements to zero.

\textbf{Shrinkage estimator}: Shrinkage methods perform a convex combination
of a singular matrix $\Sigma_m$ and some stable \emph{target} matrix. The
Shrinkage estimator~\citep{Shrinkage} $\Sigma_{\text{Shrinkage}}$ employs a
diagonal target matrix, where elements on the diagonal equal to the mean of
$\Sigma_m$ eigenvalues.

\textbf{Ledoit-Wolf estimator}: \citet{Ledoit} proposed a method to compute
the shrinkage coefficient $\hat{\alpha}$, that minimizes the expected mean square
error between $\Sigma_{\text{Shrinkage}}$ and the unobserved true covariance
matrix $\Sigma^{*}$. We refer the interested reader to \citet{Ledoit,honey}
for the exact expression for $\hat{\alpha}$ and its derivation.

\begin{table}[H]
	\centering
	\begin{adjustbox}{width=0.49\textwidth}
		$\begin{array}{ ll }
			\toprule
			\text{Estimator} & \text{Formula} \\
			\midrule
			\text{Maximum Likelihood} & \Sigma_m=\frac{1}{N}\sum\limits_{i=1}^{N}(x_i
			- \bar{x})(x_i-\bar{x})^T \\
			\addlinespace[0.2cm]
			\text{Van Ness} & \Sigma_{\text{Van Ness}} = \beta \mathop{\mathrm{diag}}(\Sigma_m) \\
			\addlinespace[0.2cm]
			\text{Shrinkage} & \Sigma_{\text{Shrinkage}}=(1 - \alpha) \Sigma_m
			+ \alpha \frac{Tr(\Sigma_m)}{d} \mathbb{I} \\
			\addlinespace[0.2cm]
			\text{Ledoit-Wolf} & \Sigma_{\text{Ledoit-Wolf}}=(1 - \hat{\alpha})\Sigma_m
			+ \hat{\alpha} \frac{Tr(\Sigma_m)}{d} \mathbb{I} \\
			\bottomrule
		\end{array}$
	\end{adjustbox}
	\caption{Robust covariance estimators. $\beta \in R$ and $\alpha \in (0,
      1)$ are the hyper-parameters of these estimators.}
	\label{tab:estimators}
\end{table}

\section{Experiments}
\label{sec:experiments}

Low sample covariance matrix estimators $\Sigma_{\text{Van Ness}}$,
$\Sigma_{\text{Shrinkage}}$, and $\Sigma_{\text{Ledoit-Wolf}}$ can be used as
drop-in replacements for MLE $\Sigma_{\text{m}}$ in the Mahalanobis
distance. We compare Mahalanobis distance-based OOD detection with those 4
alternative covariance estimation methods. Additionally, we benchmark the
candidate methods against the energy-based OOD detection~\citep{EnergyOOD},
and the gradient norm approach~\citep{GradNorm}. Finally, we compare with the
Maximum Softmax Probability (MSP) approach~\citep{hendrycksMSP}, which was
shown to be a strong baseline for OOD detection.

\begin{table*}[t!]
	\centering
	\resizebox{\textwidth}{!}{%
				\begin{tabular}{@{}lllll@{}}
			\toprule
			Dataset (\textbf{5-shot}) &       OOD method             & AUC $\uparrow$                          & PR ROC $_{ood\_pos}$ $\uparrow$ & FPR $_{ood\_pos}$ $\downarrow$ \\ \midrule
			ROSTD & MLE          & 85.13$\pm$5.41          & 58.49$\pm$10.74         & 36.16$\pm$9.47          \\
			& Van Ness           & 93.87$\pm$3.41          & 81.70$\pm$9.06          & 20.45$\pm$9.70          \\
			& Shrinkage          & \textbf{94.39$\pm$2.88} & \textbf{83.32$\pm$7.79}         & \textbf{19.25$\pm$9.09}        \\
			& Ledoit-wolf        & 94.33$\pm$2.95          & 83.19$\pm$7.90          & 19.40$\pm$9.14          \\
			& Grad Norm          & 93.47$\pm$3.32          & 81.04$\pm$8.76          & 22.68$\pm$10.06         \\
			& Energy             & 93.92$\pm$3.04          & 81.20$\pm$9.13          & 20.01$\pm$9.03          \\
			& MSP                & 92.13$\pm$3.41          & 78.23$\pm$8.34          & 25.69$\pm$9.07          \\ \midrule
			SNIPS & MLE          & 76.63$\pm$9.47          & 49.68$\pm$13.00         & 56.53$\pm$11.65         \\
			& Van Ness           & \textbf{90.46$\pm$3.31}                 & \textbf{73.98$\pm$8.21}         & \textbf{29.77$\pm$8.35}        \\
			& Shrinkage          & 90.23$\pm$3.33          & 73.35$\pm$8.17          & 30.00$\pm$8.40          \\
			& Ledoit-wolf        & 90.23$\pm$3.34          & 73.38$\pm$8.18          & 29.91$\pm$8.42          \\
			& Grad Norm          & 88.87$\pm$4.76          & 72.04$\pm$10.92         & 37.29$\pm$15.39         \\
			& Energy             & 88.92$\pm$5.74          & 70.32$\pm$11.53         & 33.30$\pm$14.38         \\
			& MSP                & 89.27$\pm$4.05          & 71.71$\pm$8.46          & 33.14$\pm$11.13         \\ \midrule
			ROSTD Coarse     & MLE & 71.39$\pm$8.70                          & 40.68$\pm$9.97                  & 65.87$\pm$12.45                \\
			& Van Ness           & 94.79$\pm$2.71          & 82.83$\pm$8.78          & 16.06$\pm$6.93          \\
			& Shrinkage          & \textbf{94.94$\pm$2.85}                 & \textbf{83.85$\pm$8.76}         & 15.92$\pm$6.88       \\
			& Ledoit-wolf        & 94.93$\pm$2.85          & 83.83$\pm$8.78          & \textbf{15.91$\pm$6.84}          \\
			& Grad Norm          & 92.38$\pm$3.59          & 77.69$\pm$9.90          & 23.84$\pm$10.32         \\
			& Energy             & 93.43$\pm$3.02          & 78.65$\pm$9.55          & 19.66$\pm$7.89          \\
			& MSP                & 92.96$\pm$2.78          & 77.05$\pm$8.92          & 19.54$\pm$6.46          \\ \midrule
			Clinic 150       & MLE & 94.34$\pm$0.45                          & 78.23$\pm$2.07                  & 23.05$\pm$2.02                 \\
			& Van Ness           & 94.42$\pm$0.39          & 79.11$\pm$1.60          & 23.08$\pm$2.00          \\
			& Shrinkage          & 94.84$\pm$0.41          & 81.44$\pm$1.83          & 21.90$\pm$1.88          \\
			& Ledoit-wolf        & 94.76$\pm$0.42          & 81.07$\pm$1.78          & 22.29$\pm$1.93          \\
			& Grad Norm          & 94.94$\pm$0.40          & 81.48$\pm$1.67          & \textbf{21.67$\pm$2.21} \\
			& Energy             & \textbf{94.95$\pm$0.38} & \textbf{81.56$\pm$1.67} & 21.85$\pm$2.23          \\
			& MSP                & 94.08$\pm$0.42          & 78.41$\pm$1.64          & 25.21$\pm$1.88          \\ \bottomrule
		\end{tabular}%
			    \hspace{8em}

		\begin{tabular}{@{}lllll@{}}
			\toprule
			Dataset (\textbf{10-shot}) &      OOD method     & AUC $\uparrow$                           & PR ROC $_{ood\_pos}$ $\uparrow$ & FPR $_{ood\_pos}$ $\downarrow$ \\ \midrule
			ROSTD FINE        & MLE & \multicolumn{1}{r}{94.44$\pm$2.45}          & 80.30$\pm$8.35                     & 14.98$\pm$5.45                    \\
			& Van Ness    & \multicolumn{1}{r}{96.79$\pm$1.43} & 89.04$\pm$5.33          & 10.26$\pm$3.25          \\
			& Shrinkage   & \multicolumn{1}{r}{\textbf{97.37$\pm$1.23}} & \textbf{90.85$\pm$4.94}            & \textbf{8.52$\pm$2.96}            \\
			& Ledoit-wolf & \multicolumn{1}{r}{97.27$\pm$1.27} & 90.51$\pm$5.05          & 8.78$\pm$3.02           \\
			& Grad Norm   & \multicolumn{1}{r}{96.17$\pm$1.42} & 88.14$\pm$4.42          & 13.33$\pm$4.15          \\
			& Energy      & \multicolumn{1}{r}{96.98$\pm$1.23} & 89.83$\pm$4.49          & 10.10$\pm$3.47          \\
			& MSP         & \multicolumn{1}{r}{95.32$\pm$1.55} & 84.32$\pm$5.49          & 13.39$\pm$3.15          \\ \midrule
			SNIPS        & MLE    & \multicolumn{1}{r}{85.27$\pm$9.24} & 62.58$\pm$13.87         & 40.23$\pm$21.25         \\
			& Van Ness    & \multicolumn{1}{r}{90.57$\pm$4.45} & 74.35$\pm$10.99         & 28.81$\pm$9.62          \\
			& Shrinkage   & \multicolumn{1}{r}{\textbf{90.68$\pm$4.51}} & \textbf{74.79$\pm$11.21}           & \textbf{28.44$\pm$9.77}           \\
			& Ledoit-wolf & \multicolumn{1}{r}{90.67$\pm$4.51} & 74.70$\pm$11.19         & 28.48$\pm$9.49          \\
			& Grad Norm   & \multicolumn{1}{r}{85.73$\pm$7.27} & 68.93$\pm$12.46         & 51.42$\pm$19.06         \\
			& Energy      & \multicolumn{1}{r}{89.75$\pm$5.48} & 73.57$\pm$12.16         & 32.89$\pm$15.11         \\
			& MSP         & \multicolumn{1}{r}{89.27$\pm$4.61} & 70.81$\pm$10.67         & 32.63$\pm$13.17         \\ \midrule
			ROSTD COARSE & MLE    & 87.23$\pm$8.20                     & 64.96$\pm$11.07         & 35.16$\pm$20.79         \\
			& Van Ness    & 95.99$\pm$1.87                     & 85.98$\pm$6.21          & 11.40$\pm$4.76          \\
			& Shrinkage   & \textbf{96.48$\pm$1.82}            & \textbf{87.96$\pm$5.92} & \textbf{10.26$\pm$4.83} \\
			& Ledoit-wolf & 96.45$\pm$1.83                     & 87.86$\pm$5.92          & 10.34$\pm$4.83          \\
			& Grad Norm   & 93.38$\pm$3.02                     & 81.31$\pm$7.46          & 23.97$\pm$13.61         \\
			& Energy      & 95.35$\pm$2.07                     & 84.01$\pm$8.36          & 13.99$\pm$5.55          \\
			& MSP         & 94.58$\pm$2.21                     & 81.00$\pm$7.22          & 14.53$\pm$5.27          \\ \midrule
			CLINIC 150   & MLE    & \textbf{96.07$\pm$0.25}            & 85.54$\pm$1.20          & \textbf{18.03$\pm$1.10} \\
			& Van Ness    & 95.73$\pm$0.26                     & 83.96$\pm$1.10          & 18.94$\pm$0.96          \\
			& Shrinkage   & 96.05$\pm$0.24                     & \textbf{85.55$\pm$1.11} & 18.12$\pm$1.04          \\
			& Ledoit-wolf & 96.00$\pm$0.24                     & 85.37$\pm$1.07          & 18.30$\pm$1.05          \\
			& Grad Norm   & 95.81$\pm$0.27                     & 85.30$\pm$1.01          & 18.70$\pm$1.19          \\
			& Energy      & 95.99$\pm$0.24                     & 85.34$\pm$1.03          & 18.54$\pm$1.09          \\
			& MSP         & 95.25$\pm$0.28                     & 82.59$\pm$1.09          & 20.05$\pm$1.19          \\ \bottomrule
		\end{tabular}%
	}
	\caption{Comparison of covariance-corrected Mahalanobis distance methods
      for OOD detection with the baselines in \textbf{5-shot} and
      \textbf{10-shot} settings.}
	\label{tab:10shot_result}
\end{table*}

\subsection{Training Procedure}
\label{sec:train_procedure}

In all our experiments, we fine-tune the RoBERTa~\citep{roberta} base model
for intent classification using the cross-entropy loss. We start from model
weights which were pre-trained on five English-language corpora of varying
sizes and domains, totaling over 160GB of uncompressed text.

Fine-tuning was performed using the AdamW~\citep{loshchilov2018decoupled}
optimizer with learning rate of $2e^{-5}$ and linear decay for 45 epochs. We
repeat all experiments 20 times, sampling a new training set at each
iteration. For hyperparameters, we use a fixed shrinkage coefficient $\alpha =
0.1$, and a fixed Van Ness hyperparameter $\beta = 1.0$ in all the
experiments. As suggested in \citet{EnergyOOD,GradNorm}, we set the
temperature $\tau = 1$ for the energy and gradient norm baselines.

During $X$-shot training, covariance matrix of size $d \times d$ is computed
from a data matrix of size $(X \cdot N_c) \times d$, where $d = 768$ for
RoBERTa embeddings and $N_c$ is the number of classes. On each iteration of
the experiment, 1) we randomly sample $X$ data points per class from the
training set, and 2) compute covariance matrix using only $X \cdot N_c$ data
points using corresponding estimators.

\subsection{Datasets and Metrics}
\label{sec:datasets}

We evaluate our approach on the following datasets:
\begin{enumerate}
	\item \textbf{CLINC150}~\citep{clinic150}: was proposed to evaluate the
	performance of task-oriented dialogue systems on out-of-domain queries. The
	dataset contains 150 intents spanning over 10 domains.

	\item \textbf{ROSTD}~\citep{originalrostd}: was developed to test
	cross-lingual transfer learning for multilingual task oriented dialog. The
	dataset was later extended by \citet{gangal} by adding OOD intents to
	English language utterances.

	\item \textbf{ROSTD-COARSE}: following \citet{Podolskiy2021RevisitingMD}, we
	also experiment with a coarsened version of ROSTD with only 3 intent
	classes. We use the same set of OOD intents for testing for both
	fine-grained and coarsened version of the dataset.

	\item \textbf{SNIPS}~\citep{snips}: contains 7 intents with approximately 2000
	utterances per intent. Since the dataset does not provide IND and OOD split,
	we follow the protocol from \citet{Podolskiy2021RevisitingMD} by randomly
	taking 5 intents as IND and the remaining 2 intents as OOD. We sample
	different OOD and IND intents on each iteration of the experiment.
\end{enumerate}

Since OOD detection is framed is a binary classification problem, we evaluate
performance in terms of AUC, PR ROC, and FPR@95\%TPR metrics. For FPR, the
decision threshold is chosen so that the True Positive Rate is 95\%. We omit
95\% suffix from the metric names to avoid repetition in notation. We report
measurements for the cases when the OOD class is treated as positive, and when
the OOD class is treated as negative. This is indicated with ${ood\_pos}$ and
${ood_\_neg}$ suffixes correspondingly.

\section{Results}
\label{sec:results}
Table~\ref{tab:10shot_result} compares the performance of different variants
of the Mahalanobis distance on benchmarking datasets. We observe that
covariance corrected methods (Shrinkage, Ledoit-Wolf and Van Ness) outperform
other OOD detection techniques on 3 out of 4 datasets on all metrics of
interest. The table is summarized in Figure~\ref{fig:baseline_comparison} for AUC
using the MLE and Shrinkage estimators.

Furthermore, in 5-shot setting the covariance correction outperforms MLE on
all datasets. However, MLE performs competitively with covariance correction
when the number of samples used for covariance estimation increases. This
phenomenon can be seen for the CLINC150 dataset in 10-shot setting, where the
$768 \times 768$ dimensional covariance matrix was computed by using $150
\cdot 10 = 1500$ samples (150 classes and 10 samples per class).

\begin{figure}[t]
	   \centering
		\includegraphics[width=0.9\columnwidth]{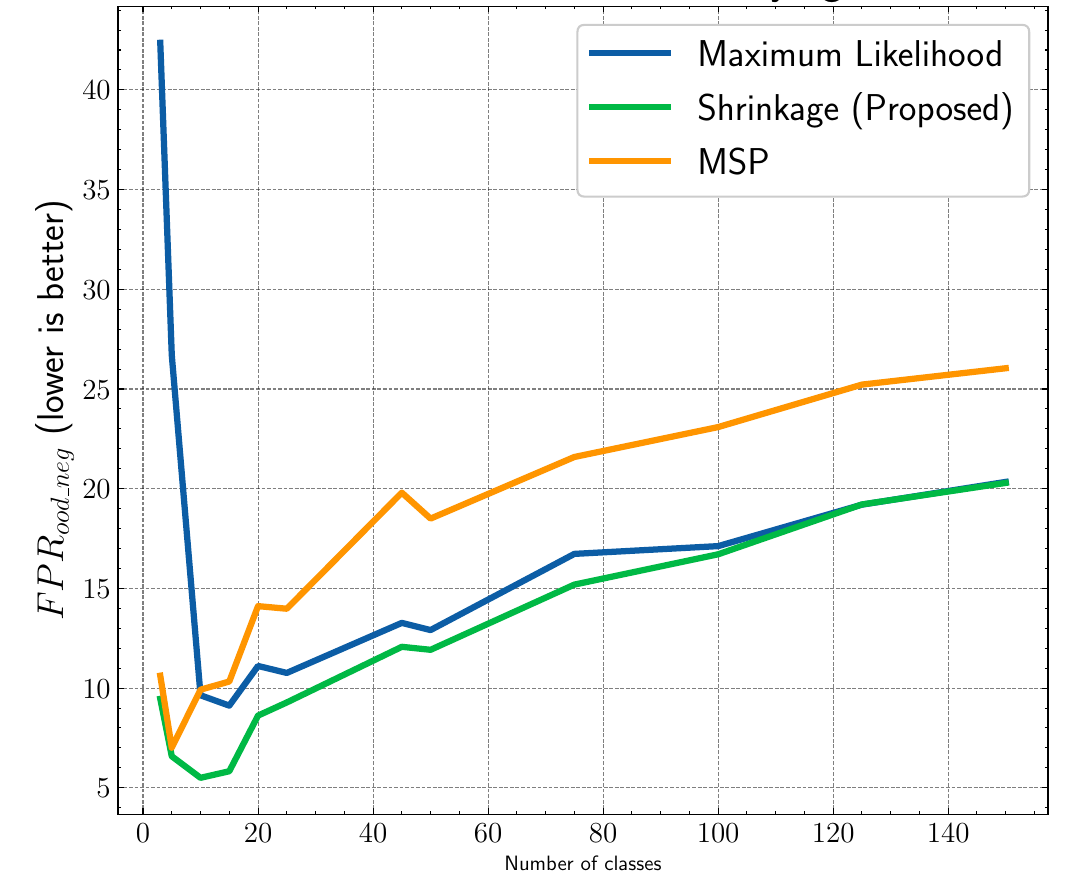}
		\caption{Performance on the CLINC150 dataset with respect to the number of
		training classes in \textbf{10-shot} setting.}
		\label{fig:varying_class}
\end{figure}

To assess the impact of OOD performance on the number of intent classes, we
train several models by sampling subset of intents from the CLINC150
dataset. In Figure~\ref{fig:varying_class} we observe that Shrinkage estimator
outperforms MLE by a large margin, when the number of intents is small and it
converges to MLE performance as the number of classes increase. The
effectiveness of OOD detection degrades as the number of in-domain (IND)
classes increases, which is yet another interesting phenomenon. The reason is
that, when the number of OOD samples are fixed, an increase in the size of IND
samples leads to greater confusion with OOD class, because more samples end up
on the IND/OOD boundary. Quantitatively, going from 3 to 20 IND classes causes
the minimum distance from a query OOD example to the closest IND centroid
decreases by 20\%, this confuses the methods that rely on a fixed score
threshold. When all 150 classes are used then, the average IND-OOD Euclidean
distance decreases by 37\%. Moreover, as the amount of data increases, the
performance of the robust estimators degrades, but at different rates.

\section{Conclusion}
\label{sec:conclusion}
We have demonstrated that in few-shot settings Mahalanobis distance computed
using robust covariance estimators consistently outperforms the Maximum
Likelihood estimator baseline. According to our experiments, the Shrinkage
estimator excels in 5-shot and 10-shot settings across various datasets. The
suggested approach is computationally cheap, with an additional overhead of
one matrix-vector multiplication operation per class, and does not require any
auxiliary data nor modifications to the training procedure.

\bibliography{aaai23}

@inproceedings{Podolskiy2021RevisitingMD,
	title={Revisiting Mahalanobis Distance for Transformer-Based Out-of-Domain Detection},
	author={A. V. Podolskiy and Dmitry Lipin and A. Bout and E. Artemova and Irina Piontkovskaya},
	booktitle={AAAI},
	year={2021}
}

@inproceedings{loshchilov2018decoupled,
	title={Decoupled Weight Decay Regularization},
	author={Loshchilov, Ilya and Hutter, Frank},
	booktitle={International Conference on Learning Representations},
	year={2018}
}

@article{surveyIntent,
	author = {Chen, Hongshen and Liu, Xiaorui and Yin, Dawei and Tang, Jiliang},
	title = {A Survey on Dialogue Systems: Recent Advances and New Frontiers},
	year = {2017},
	issue_date = {December 2017},
	publisher = {Association for Computing Machinery},
	address = {New York, NY, USA},
	volume = {19},
	number = {2},
	issn = {1931-0145},
	url = {https://doi.org/10.1145/3166054.3166058},
	doi = {10.1145/3166054.3166058},
	journal = {SIGKDD Explor. Newsl.},
	month = {nov},
	pages = {25–35},
	numpages = {11}
}

@article{Hendrycks2019DeepAD,
	title={Deep Anomaly Detection with Outlier Exposure},
	author={Dan Hendrycks and Mantas Mazeika and Thomas G. Dietterich},
	journal={ArXiv},
	year={2019},
	volume={abs/1812.04606}
}

@article{pnPOOD,
	author    = {Mrinal Rawat and
	Ramya Hebbalaguppe and
	Lovekesh Vig},
	title     = {PnPOOD : Out-Of-Distribution Detection for Text Classification via
	Plug andPlay Data Augmentation},
	journal   = {CoRR},
	volume    = {abs/2111.00506},
	year      = {2021},
	url       = {https://arxiv.org/abs/2111.00506},
	eprinttype = {arXiv},
	eprint    = {2111.00506},
	bibsource = {dblp computer science bibliography, https://dblp.org}
}

@article{goldpaper,
	author    = {Derek Chen and
	Zhou Yu},
	title     = {{GOLD:} Improving Out-of-Scope Detection in Dialogues using Data Augmentation},
	journal   = {CoRR},
	volume    = {abs/2109.03079},
	year      = {2021},
	url       = {https://arxiv.org/abs/2109.03079},
	eprinttype = {arXiv},
	eprint    = {2109.03079},
	bibsource = {dblp computer science bibliography, https://dblp.org}
}

@inproceedings{MahalanobisOrig,
	author = {Lee, Kimin and Lee, Kibok and Lee, Honglak and Shin, Jinwoo},
	title = {A Simple Unified Framework for Detecting Out-of-Distribution Samples and Adversarial Attacks},
	year = {2018},
	publisher = {Curran Associates Inc.},
	address = {Red Hook, NY, USA},
	booktitle = {Proceedings of the 32nd International Conference on Neural Information Processing Systems},
	pages = {7167–7177},
	numpages = {11},
	location = {Montr\'{e}al, Canada},
	series = {NIPS'18}
}

@article{EnergyOOD,
	author    = {Weitang Liu and
	Xiaoyun Wang and
	John D. Owens and
	Yixuan Li},
	title     = {Energy-based Out-of-distribution Detection},
	journal   = {CoRR},
	volume    = {abs/2010.03759},
	year      = {2020},
	url       = {https://arxiv.org/abs/2010.03759},
	eprinttype = {arXiv},
	eprint    = {2010.03759},
	bibsource = {dblp computer science bibliography, https://dblp.org}
}

@article{GradNorm,
	author    = {Rui Huang and
	Andrew Geng and
	Yixuan Li},
	title     = {On the Importance of Gradients for Detecting Distributional Shifts
	in the Wild},
	journal   = {CoRR},
	volume    = {abs/2110.00218},
	year      = {2021},
	url       = {https://arxiv.org/abs/2110.00218},
	eprinttype = {arXiv},
	eprint    = {2110.00218},
	bibsource = {dblp computer science bibliography, https://dblp.org}
}

@article{ODIN,
	author    = {Shiyu Liang and
	Yixuan Li and
	R. Srikant},
	title     = {Principled Detection of Out-of-Distribution Examples in Neural Networks},
	journal   = {CoRR},
	volume    = {abs/1706.02690},
	year      = {2017},
	url       = {http://arxiv.org/abs/1706.02690},
	eprinttype = {arXiv},
	eprint    = {1706.02690},
	bibsource = {dblp computer science bibliography, https://dblp.org}
}

@article{hendrycksMSP,
	author    = {Dan Hendrycks and
	Kevin Gimpel},
	title     = {A Baseline for Detecting Misclassified and Out-of-Distribution Examples
	in Neural Networks},
	journal   = {CoRR},
	volume    = {abs/1610.02136},
	year      = {2016},
	url       = {http://arxiv.org/abs/1610.02136},
	eprinttype = {arXiv},
	eprint    = {1610.02136},
	bibsource = {dblp computer science bibliography, https://dblp.org}
}

@article{mahalanobisSurvey,
	title={The mahalanobis distance},
	author={De Maesschalck, Roy and Jouan-Rimbaud, Delphine and Massart, D{\'e}sir{\'e} L},
	journal={Chemometrics and intelligent laboratory systems},
	volume={50},
	number={1},
	pages={1--18},
	year={2000},
	publisher={Elsevier}
}

@article{anomalydetection,
	title={A low-rank and sparse matrix decomposition-based Mahalanobis distance method for hyperspectral anomaly detection},
	author={Zhang, Yuxiang and Du, Bo and Zhang, Liangpei and Wang, Shugen},
	journal={IEEE Transactions on Geoscience and Remote Sensing},
	volume={54},
	number={3},
	pages={1376--1389},
	year={2015},
	publisher={IEEE}
}

@article{NoClearWinner,
	author    = {Fahim Tajwar and
	Ananya Kumar and
	Sang Michael Xie and
	Percy Liang},
	title     = {No True State-of-the-Art? {OOD} Detection Methods are Inconsistent
	across Datasets},
	journal   = {CoRR},
	volume    = {abs/2109.05554},
	year      = {2021},
	url       = {https://arxiv.org/abs/2109.05554},
	eprinttype = {arXiv},
	eprint    = {2109.05554},
	bibsource = {dblp computer science bibliography, https://dblp.org}
}

@article{vanness,
	author    = {John Van Ness},
	title     = {On the dominance of non-parametric Bayes rule discriminant algorithms
	in high dimensions},
	journal   = {Pattern Recognit.},
	volume    = {12},
	number    = {6},
	pages     = {355--368},
	year      = {1980},
	url       = {https://doi.org/10.1016/0031-3203(80)90012-6},
	doi       = {10.1016/0031-3203(80)90012-6},
	bibsource = {dblp computer science bibliography, https://dblp.org}
}

@article{Shrinkage,
	title={Regularized Discriminant Analysis},
	author={Jerome H. Friedman},
	journal={Journal of the American Statistical Association},
	year={1989},
	volume={84},
	pages={165-175}
}

@article{Ledoit,
	title={A well-conditioned estimator for large-dimensional covariance matrices},
	author={Olivier Ledoit and Michael Wolf},
	journal={Journal of Multivariate Analysis},
	year={2004},
	volume={88},
	pages={365-411}
}

@article{honey,
	author = {Ledoit, Olivier and Wolf, Michael},
	year = {2003},
	month = {07},
	pages = {},
	title = {Honey, I Shrunk the Sample Covariance Matrix},
	volume = {30},
	journal = {The Journal of Portfolio Management},
	doi = {10.2139/ssrn.433840}
}

@inproceedings{clinic150,
	title = "An Evaluation Dataset for Intent Classification and Out-of-Scope Prediction",
	author = "Larson, Stefan  and
	Mahendran, Anish  and
	Peper, Joseph J.  and
	Clarke, Christopher  and
	Lee, Andrew  and
	Hill, Parker  and
	Kummerfeld, Jonathan K.  and
	Leach, Kevin  and
	Laurenzano, Michael A.  and
	Tang, Lingjia  and
	Mars, Jason",
	booktitle = "Proceedings of the 2019 Conference on Empirical Methods in Natural Language Processing and the 9th International Joint Conference on Natural Language Processing (EMNLP-IJCNLP)",
	month = nov,
	year = "2019",
	address = "Hong Kong, China",
	publisher = "Association for Computational Linguistics",
	url = "https://aclanthology.org/D19-1131",
	doi = "10.18653/v1/D19-1131",
	pages = "1311--1316",
}

@article{originalrostd,
	author    = {Sebastian Schuster and
	Sonal Gupta and
	Rushin Shah and
	Mike Lewis},
	title     = {Cross-lingual Transfer Learning for Multilingual Task Oriented Dialog},
	journal   = {CoRR},
	volume    = {abs/1810.13327},
	year      = {2018},
	url       = {http://arxiv.org/abs/1810.13327},
	eprinttype = {arXiv},
	eprint    = {1810.13327},
	bibsource = {dblp computer science bibliography, https://dblp.org}
}

@article{gangal,
	author    = {Varun Gangal and
	Abhinav Arora and
	Arash Einolghozati and
	Sonal Gupta},
	title     = {Likelihood Ratios and Generative Classifiers for Unsupervised Out-of-Domain
	Detection In Task Oriented Dialog},
	journal   = {CoRR},
	volume    = {abs/1912.12800},
	year      = {2019},
	url       = {http://arxiv.org/abs/1912.12800},
	eprinttype = {arXiv},
	eprint    = {1912.12800},
	bibsource = {dblp computer science bibliography, https://dblp.org}
}

@article{snips,
	title   = {Snips Voice Platform: an embedded Spoken Language Understanding system for private-by-design voice interfaces},
	author  = {Coucke, Alice and Saade, Alaa and Ball, Adrien and Bluche, Th{\'e}odore and Caulier, Alexandre and Leroy, David and Doumouro, Cl{\'e}ment and Gisselbrecht, Thibault and Caltagirone, Francesco and Lavril, Thibaut and others},
	journal = {arXiv preprint arXiv:1805.10190},
	pages   = {12--16},
	year    = {2018}
}

@misc{roberta,
	doi = {10.48550/ARXIV.1907.11692},
	
	url = {https://arxiv.org/abs/1907.11692},
	
	author = {Liu, Yinhan and Ott, Myle and Goyal, Naman and Du, Jingfei and Joshi, Mandar and Chen, Danqi and Levy, Omer and Lewis, Mike and Zettlemoyer, Luke and Stoyanov, Veselin},
	
	title = {RoBERTa: A Robustly Optimized BERT Pretraining Approach},
	
	publisher = {arXiv},
	
	year = {2019},
	
	copyright = {arXiv.org perpetual, non-exclusive license}
}

\end{document}